\UseRawInputEncoding
\documentclass[conference]{ieeeconf}
\usepackage{amsmath}
\usepackage{xcolor, soul}
\sethlcolor{yellow}
\usepackage{threeparttable}
\usepackage{multirow}
\usepackage{graphicx}
\usepackage{algorithm}
\usepackage{algpseudocode}
\usepackage{subfigure}
\usepackage{tabularx}
\usepackage{xcolor}
\usepackage{booktabs}
\usepackage{balance}
\usepackage{siunitx}
\usepackage{layouts}
\usepackage{cite}

\makeatletter
\let\NAT@parse\undefined
\makeatother

\graphicspath{{../}}

\begin{document}
\title{SoGraB: A Visual Method for Soft Grasping Benchmarking and Evaluation}

\author {Benjamin G. Greenland$^{1,2}$, Josh Pinskier$^{1}$, Xing Wang$^{1,*}$, Daniel Nguyen$^{3}$, Ge Shi$^{1}$, \\Tirthankar Bandyopadhyay$^{1}$, Jen Jen Chung$^{2}$, David Howard$^{1}$}

\maketitle

\begin{abstract}

    Recent years have seen soft robotic grippers gain increasing attention due to their ability to robustly grasp soft and fragile objects. However, a commonly available standardised evaluation protocol has not yet been developed to assess the performance of varying soft robotic gripper designs. This work introduces a novel protocol, the Soft Grasping Benchmarking and Evaluation (SoGraB) method, to evaluate grasping quality, which quantifies object deformation by using the Density-Aware Chamfer Distance (DCD) between point clouds of soft objects before and after grasping. We validated our protocol in extensive experiments, which involved ranking three Fin-Ray gripper designs with a subset of the EGAD object dataset. The protocol appropriately ranked grippers based on object deformation information, validating the method's ability to select soft grippers for complex grasping tasks and benchmark them for comparison against future designs. 
    
\end{abstract}

\footnotetext[1]{CSIRO Robotics, Australia}
\footnotetext[2]{The University of Queensland, Australia}
\footnotetext[3]{Queensland University of Technology, Australia}

\IEEEpeerreviewmaketitle

\section{Introduction}
    Despite the proliferation of universal \cite{Pinskier2024,Ilievski2011} and bespoke \cite{Smith2023,Pinskier2024a,Pinskier2024c,Xie2024} soft grippers, there is still no common understanding about what makes a good gripper and how to assess them.
    the field has not yet developed standardised metrics or evaluation methods for assessing gripper design or grasp quality \cite{Baines2024}. To enhance the performance and intelligence of soft designs, a standardized framework is needed which assesses grasp quality on soft objects, ranks gripper performance, and can drive improvements in new designs. This is essential not only to demonstrate progress against soft robotic equivalents, but broadly across all robot designs \cite{doi:10.1126/scirobotics.abg6049}.

    Existing evaluation methods prioritise either grasp success rate, how often the object is successfully grasped and held \cite{Zimmer2019}; or retention force, the force required to pull the object free of the gripper \cite{10122060}. Both measures characterise the \textit{grasp quality}, the gripper's ability to grasp and hold objects. However, they don't consider the magnitude of forces applied to the object; its internal stresses or deformation; or the potential for damage. To meaningfully evaluate grasping with deformable objects, a broader grasp benchmarking method is needed which captures both \textit{grasp quality} and \textit{grasp safety}. For practicality, it should be able to be experimentally evaluated with commonly available equipment.

    In engineering design, a valid design is one which, during use, experiences stresses less than a safe proportion of its stress at failure (the factor of safety) \cite{young2001roark}, but this can only be evaluated in simulation. Hence, grasp force (the force exerted on the object by the gripper) is commonly used as a proxy. Whilst often convenient to measure, this approach requires either force sensors on the gripper \cite{Low2021}, or sensorised objects \cite{10538419, Junge2022}, preventing benchmarking of arbitrary gripper-object pairs.  
    \begin{figure}[t]
        \centering
        \includegraphics[width=\linewidth]{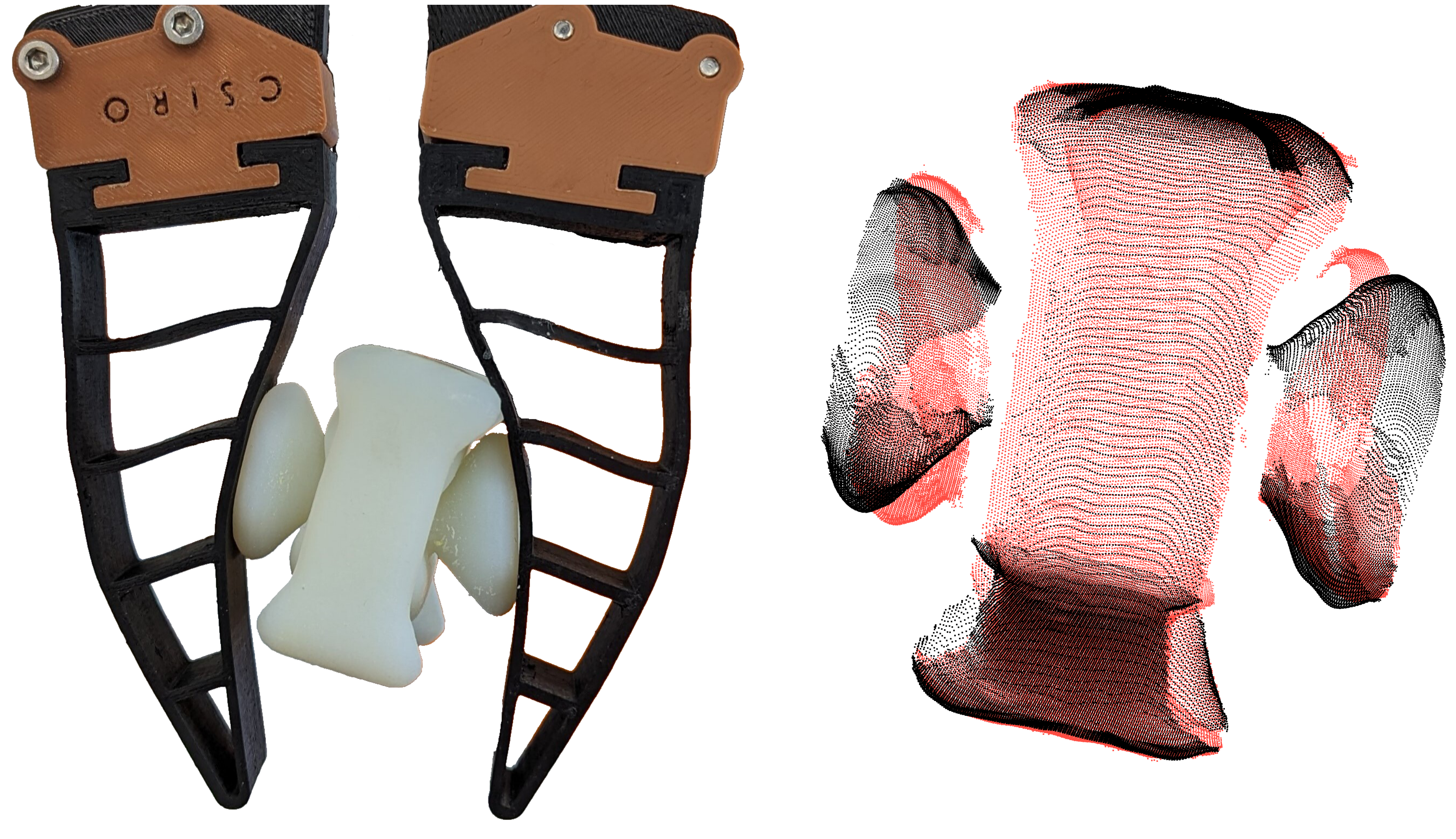}
        \caption{The SoGraB method: (Left) A soft, Shore 40A object grasped by a soft Fin-Ray gripper. (Right) The extracted point clouds showing initial state (black) and deformed state during grasping (red). By comparing the two states as well as the grasp success, we produce a grasp quality benchmark.}
        \label{fig:Heroshot}
    \end{figure}
    
    To address the urgent need to evaluate and compare different soft gripper designs, this work proposes a novel evaluation protocol, the Soft Grasping Benchmarking and Evaluation (SoGraB) method. It quantifies the grasping performance of soft grippers based on both success rate and object deformation. Deformation is used as a non-contact stress proxy and measured using 3D imaging by comparing point clouds (Figure \ref{fig:Heroshot}). 
    
    The main contributions of this paper are:
    \begin{itemize}
        \item A generalised visual methodology for soft grasp benchmarking, applicable to any object and most soft grippers 
        \item An experimental investigation of grasp quality for Fin-ray soft fingers printed at 4 hardness levels
        \item A baseline grasp dataset featuring objects of varying grasp difficulty and diverse geometries and material properties. The dataset comprises 900 grasps and their associated point clouds
    \end{itemize} 

\section{Related Work} \label{related}

\subsection{Grasp Evaluation}
    The growing uptake of robotic grasping and manipulation has generated a drive for uniform standards to assess robotic hands and end-effectors, including by the US National Institute of Standards and Technology and IEEE \cite{Falco2020}. These give a set of 11 quality metrics and associated testing procedures produced from the perspective of rigid grasping. They focus on the mechanical characteristics of the gripper rather than its effect on an object. Both the mechanical design and grasp policy are typically critical to determining grasp quality. However soft robotics blurs the distinction between the two. Soft design emphasises morphological computation, where the control policy is embodied in the gripper's mechanical design, as a surrogate for learned controllers.
    
    The basic capabilities of robotic pick and place systems are typically given by three criteria: grasp rate (picks per hour), grasp success rate (percentage of attempted picks completed) and range (a qualitative measure of the variety of objects which can be grasped) \cite{mahler_guest_2018}. While rate primarily measures the capabilities of the robot arm and grasp policy, reliability and range apply both gripper design and grasp policy and can be adapted to assess gripper designs at the object level. Further to these, grasp quality is of substantial interest, especially when grasping heterogeneous objects in unstructured environments. Grasp quality, however, is a nebulous concept and has a multitude of meanings including: disturbance rejection, grasp isotropy, area of grasped polygon, convex hull volume, hand configurations, and many more \cite{Rubert2018,roa_grasp_2015}. Even when quantitative benchmarks exist, they rely on assumptions such as the object being stationary relative to gripper during grasping \cite{pozzi_grasp_2017}. 
    
    Performance-based metrics are commonly used in real-world evaluations to provide tractable benchmarks which generalise across different hardware configurations (grippers and experimental platforms). At its simplest, it is just the grasp success rate, but quality measures such as gripper angle sensitivity \cite{sotiropoulos_benchmarking_2018},  stability (implied by holding time before an object falls) \cite{murrilo_multigrippergrasp_2024}, disturbance rejection (acceleration to dislodge an object) \cite{roa_grasp_2015}, are also used. Whilst these give a more holistic view of grasp quality than just success rate, performance-based methods rely on having standardised objects in place of standardised testing platforms to make objective comparisons. As none consider the effect of grasping on the object itself or have soft object databases, no current methods are suited to evaluating soft grippers.  

    \subsection{Evaluation Object Dataset}
    There is limited standardisation across methods for selecting objects to evaluate the performance of robotic grippers. Existing datasets typically consist of a collection of household items, like the Yale-CMU-Berkeley (YCB) \cite{calli_ycb_2015} and Dexterity Network (Dex-Net) \cite{mahler_dex-net_2016} datasets, leaving the selection of a diverse range of individual objects up to the intuition of researchers. These datasets often provide a shopping list of objects, and thus depend on the ongoing supply and local availability of products. Other datasets generate custom objects and supply their models for reproduction and replication \cite{morrison_egad_2020, 10122060}

\section{SoGraB Methodology} \label{design}
\subsection{Grasp Quality Assessment}
SoGraB evaluates soft grasping quality based on three features: grasp success, holding time, and object deformation. Together these features create a novel object-centric scalar benchmark for soft grasping quality.

Object deformation is quantified by capturing 3D point clouds of the object before and during grasping, and comparing the two  
using Density-Aware Chamfer Distance (DCD) \cite{wu_chamfer_2021}.
Unlike other common metrics like Chamfer Distance and Earth Mover's Distance, DCD is insensitive to variations in density distribution and outlying points. These features ensure more consistent and reliable evaluations when occlusions are present, making DCD suitable for comparing incomplete point clouds for deformation analysis.

The complete scoring metric is given in Equation \eqref{equ:Evaluation_Metric}, it has three scoring ranges for unsuccessful, partially successful, and successful grasps. 
An unsuccessful grasp is defined as an attempt that either failed to grasp the object or dropped the object before the grasped point cloud could be captured. 
A partially successful grasp is where the object was dropped after capturing the grasped point cloud but before it was placed down again, while a successful grasp is an attempt that held the object for a complete pick and place cycle.
\begin{equation}
        \label{equ:Evaluation_Metric}
        \text{score} = 
        \begin{cases}
            0 & \text{Unsuccessful grasp} \\
            \frac{(1-d_{\text{DCD}})t_{\text{dropped}}}{2t_{\text{cycle}}} & \text{Partially successful grasp} \\
            1 - \frac{d_{\text{DCD}}}{2} & \text{Successful grasp}
        \end{cases}
\end{equation}

The DCD algorithm is given by:
    \begin{multline}
                \label{equ:density_aware_chamfer_distance}
                d_{\text{DCD}}(S_1,S_2) = \frac{1}{2} \Bigg(\Bigg. \frac{1}{|S_1|} \sum_{x\in S_1} \left( 1-\frac{1}{n_{\hat{y}}}e^{-\alpha||x-{\hat{y}}||_2} \right) + \\
                \frac{1}{|S_2|} \sum_{y\in S_2} \left( 1-\frac{1}{n_{\hat{x}}}e^{-\alpha||y-{\hat{x}}||_2} \right) \Bigg.\Bigg)
            \end{multline}
where, $\hat{y} = min_{y \in S_2} ||x - y||_2$ and $\hat{x} = min_{x \in S_1} ||y - x||_2$. It takes the average bounded Euclidean distance between each point in one point cloud ($S_1$) and its nearest neighbour in the other point cloud ($S_2$). It bounds the distances in the range $[0, 1]$ by using the first order approximation of the Taylor Expansion ($e^z \approx 1-||x-y||_2$). A scalar, $\alpha$, is used to adjust the sensitivity of the algorithm. To ensure the algorithm is insensitive to variations in density distribution, the number of times a point is referenced as a nearest neighbour is tracked ($n_{\hat{y}}$ and $n_{\hat{x}}$), with subsequent references having a decreasing effect on the final distance. The distances are calculated with each cloud having a turn as the ``reference" cloud, with the average distance being returned, $d_{\text{DCD}}\in[0, 1]$.
    This distance was halved to differentiate between partially successful grasps, and successful grasps. Partially successful grasps were further scored on the proportion of drop time ($t_{\text{dropped}}$) compared to total cycle time ($t_{\text{cycle}}$). As a result, unsuccessful grasps are given a score of 0, partially successful grasps are scored in the range [0, 0.5] and successful grasps are scored in the range [0.5, 1].

    \subsection{Point Cloud Alignment}
\begin{figure}[t]
                \centering
                \subfigure[]{\includegraphics[width=0.35\linewidth]{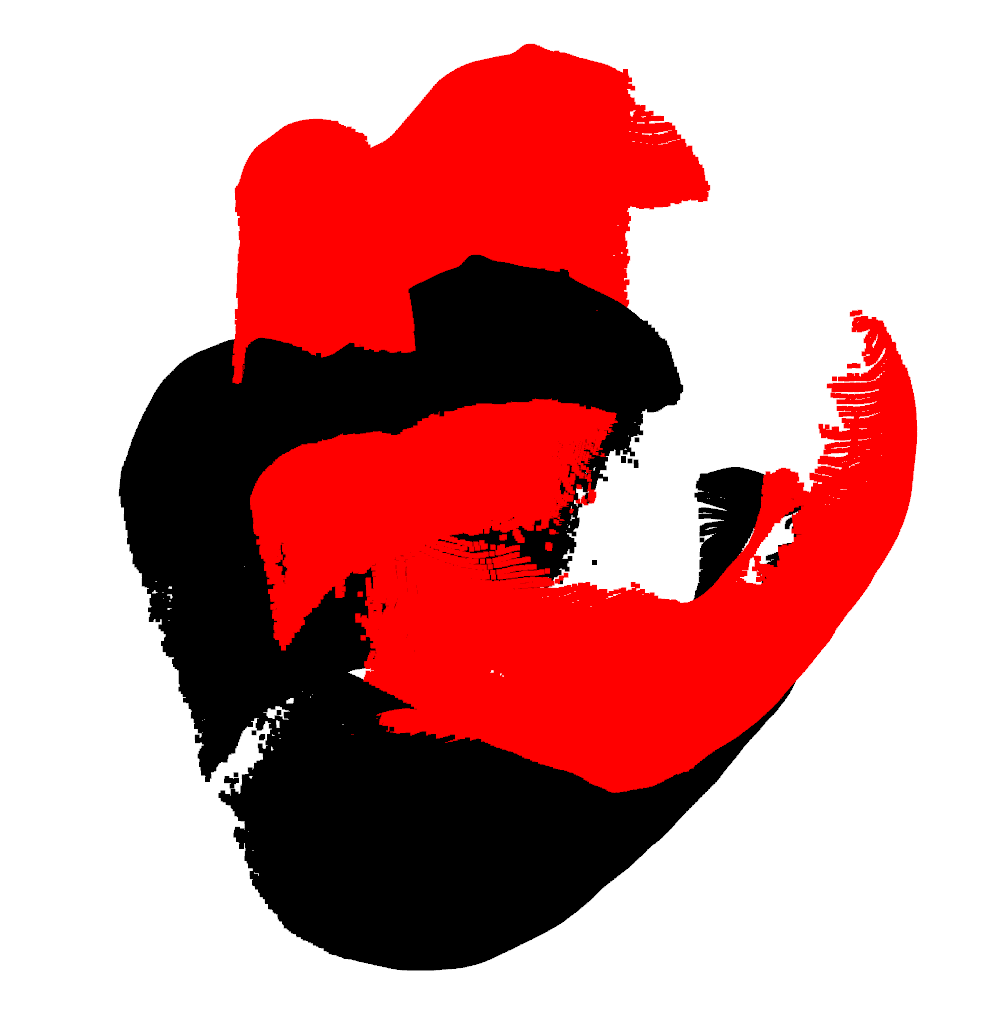}}
                \subfigure[]{\includegraphics[width=0.35\linewidth]{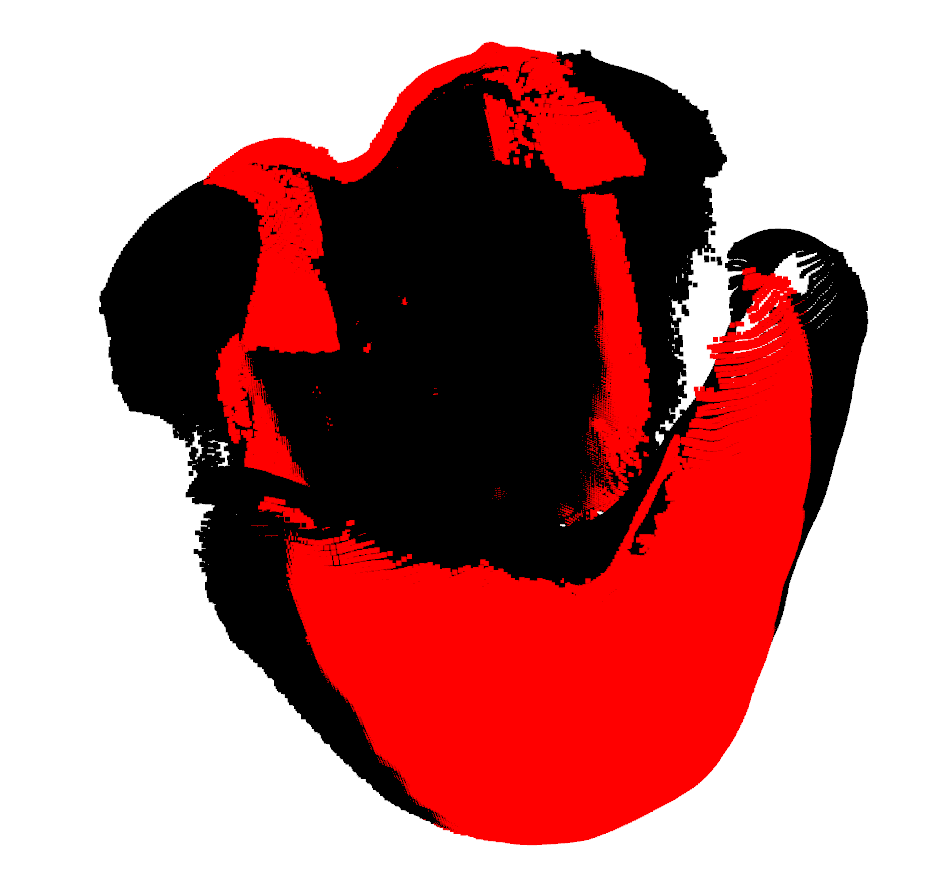}}
                \caption{Demonstration of ICP point cloud alignment (Shore 40A EGAD B1 object). (a) Initial position of point clouds, transformed into the same coordinates in the end-effector frame. (b) After ICP alignment.}
                \label{fig:ICP_vertical}
            \end{figure}

            To accurately quantify deformation and correct for any in-hand slippage or rotation (as in Figure \ref{fig:Heroshot}), the two point clouds needed to be aligned in post-processing. As the two point clouds have different sets of features (i.e. different parts of the object are visible before and after grasping) and the object changes shape during grasping, the transformation cannot be exactly calculated. We use the iterative closest point (ICP) algorithm to minimise the distance between the two point clouds. Given an approximate initial alignment taken from the robot's kinematics, this reliably approximates the true transformation for moderate deformations (Figure \ref{fig:ICP_vertical}).

            For symmetrical objects undergoing large deformations, the centre of mass and principal axes of the two point clouds were aligned. Given dense point clouds and symmetric objects, this provided a reliable transformation.

\section{Experimental Grasp Evaluation Implementation} \label{method}
\begin{figure}[t]
        \centering
        \includegraphics[width=\linewidth]{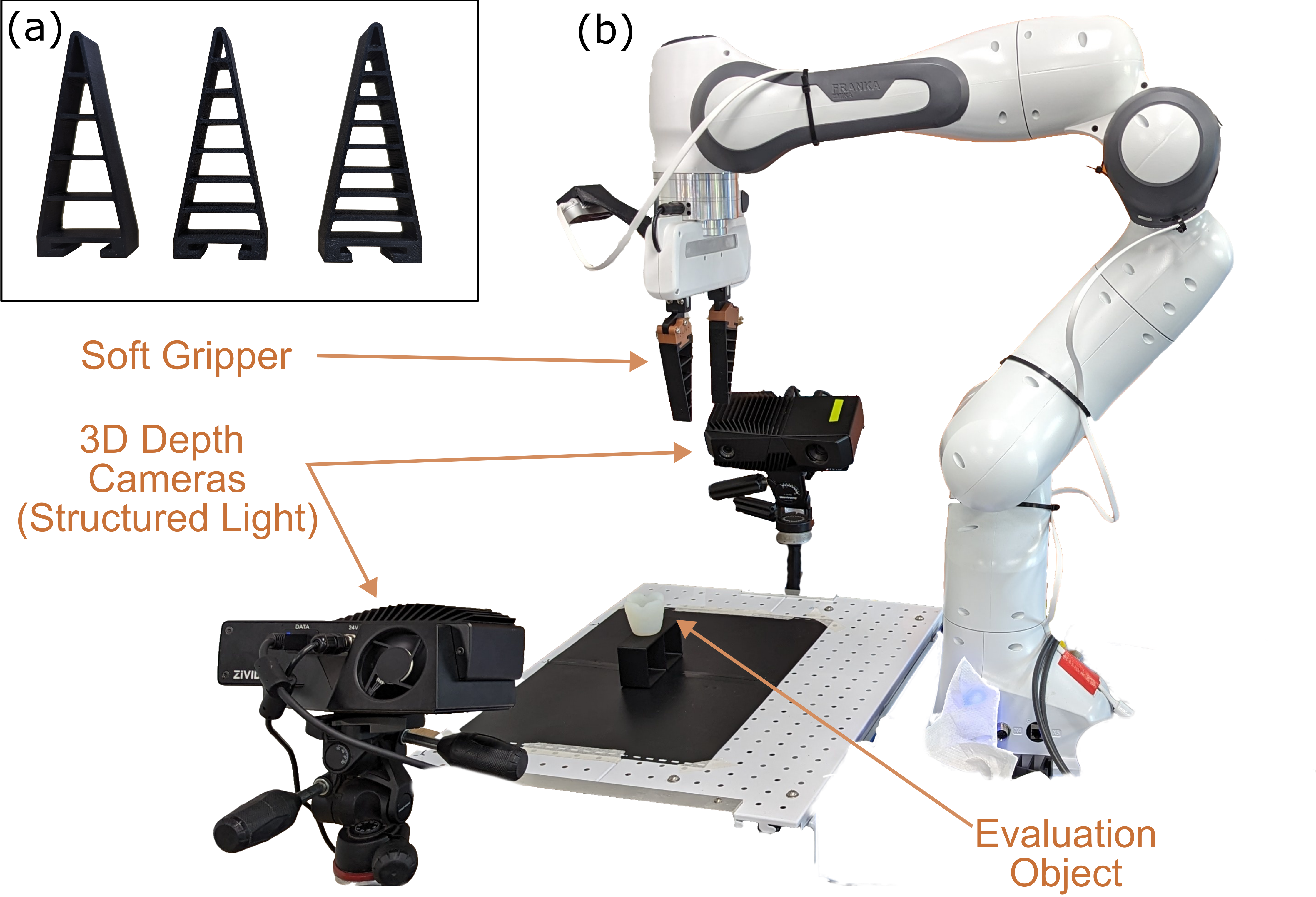}
        \caption{(a) Fin-Ray soft fingers with 4, 6 and 8 ribs. (b) Experimental grasp evaluation platform. Note: Grippers are rotated \SI{90}{\degree} in yaw prior to grasping}
        \label{fig:evaluation_setup}
    \end{figure}

    A custom experimental platform was established for the initial gripper evaluation and dataset generation (Figure \ref{fig:evaluation_setup}). The setup contained 
    a 7-DOF Franka Emika Panda manipulator and 2 Zivid One Plus Small depth cameras arranged at opposite sides of the object. The Zivid cameras use structured light to reconstruct the 3D image. To evaluate the grasp quality, the grasped objects need to be segmented out of the 3D scene and isolated from the grippers and background. To enhance this process we configure the scene to have high contrast:
    the test objects were printed in white and placed on a black background, with the test grippers also printed in black. These selections were largely dictated by the available resources at our facility, and can in principle be replaced by any similar robot arm, 3D camera and 3D printer.
    
    The test grippers were configured as an antipodal 2-finger grasp, with the arm rotated \SI{90}{\degree} about the vertical prior to grasping (as in Figure \ref{fig:Heroshot}) such that the fingers are vertical and the object is not occluded by the fingers. During testing, the gripper slowly closed around the object (\SI{0.1}{\meter\per\second}) until it applied \SI{0.5}{\newton} of force on the test object.    

    The complete procedure for evaluating a sample is:
    \begin{enumerate}
        \item Place an evaluation object in initial position and capture a point cloud of the ungrasped object.
        \item Position the gripper around the object and proceed to grasp the object (fingers vertical, with gripper plane perpendicular to cameras).
        \item Raise object and capture grasped point clouds.
        \item Raise object to manipulator's home pose (Figure \ref{fig:evaluation_setup}).
        \item Return object to the origin and release. 
    \end{enumerate}
    
    \subsection{Experimentally Evaluated Grippers}
        Fin-Ray soft fingers were selected for the initial evaluation as they are a widely used industry standard design, geometrically reconfigurable to produce different grasping stiffnesses, and simple to evaluate on standard robot arms.

        To investigate the effect of finger stiffness relative to the object, three Fin-Ray soft finger designs were evaluated, with 4, 6 and 8 internal ribs, respectively (Figure \ref{fig:evaluation_setup}(a)). All fingers were \SI{90}{\milli\meter} long, \SI{35}{\milli\meter} wide (at the base), \SI{20}{\milli\meter} thick, and have \SI{2}{\milli\meter} wide features. All were 3D printed in NinjaTek Shore 90A Eel TPU, a hardness roughly equivalent to a skateboard wheel. A fourth Fin-Ray finger was printed out of rigid PLA and served as a rigid comparison.

    \subsection{Evaluation Object Selection}
\begin{figure}[t]
            \centering
            \includegraphics[width=0.8\linewidth]{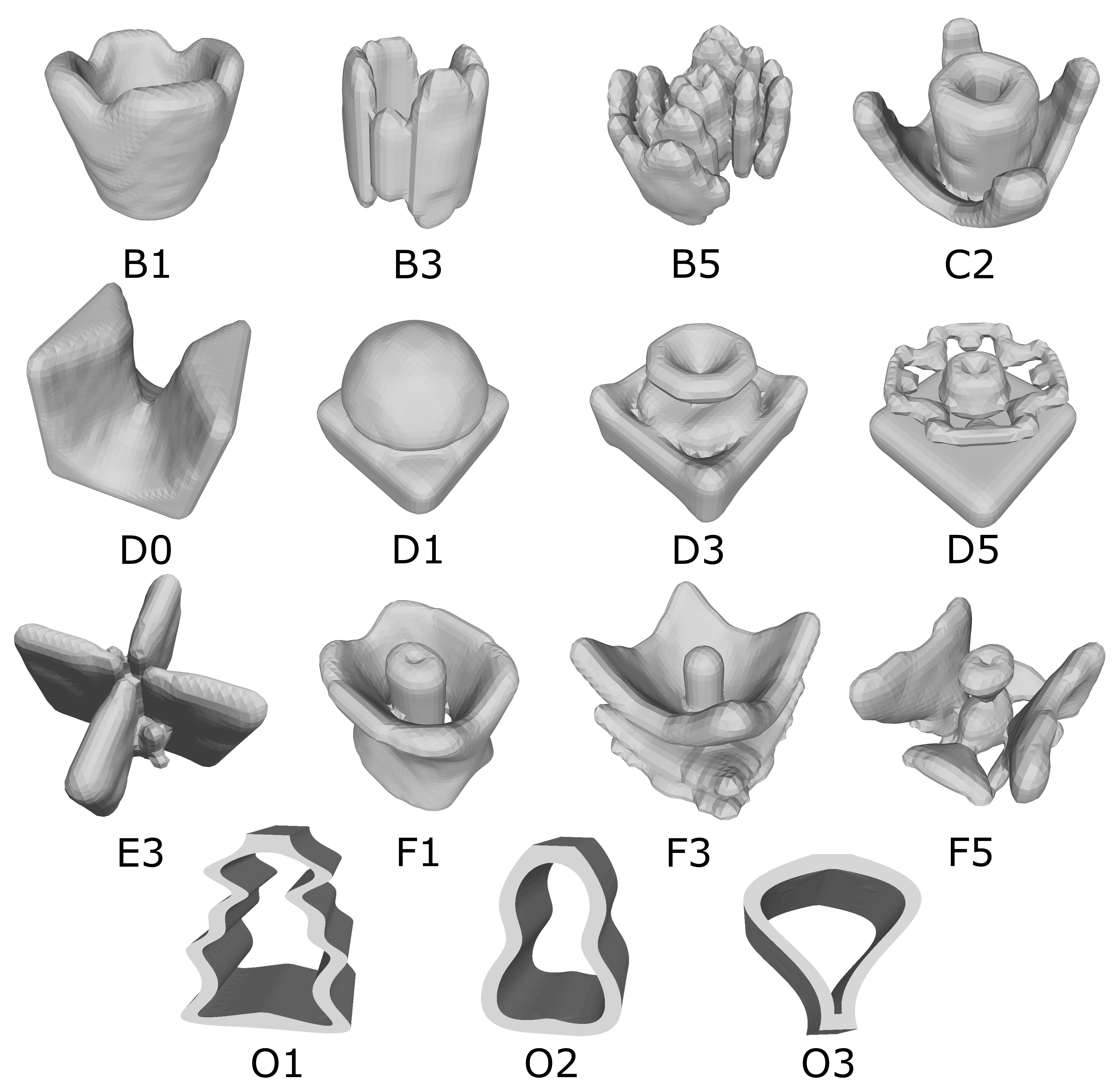}
            \caption{Evaluation objects used in this study: B1-F5 are selected objects from the EGAD dataset. O1-O3 are custom evaluation objects designed to be symmetric and highly deformable.}
            \label{fig:EGAD_subset}
        \end{figure}

        To form an initial dataset, a total of 15 objects were evaluated, comprising 12 from the EGAD evaluation dataset \cite{morrison_egad_2020}, and 3 custom generated soft objects (Figure \ref{fig:EGAD_subset}). 
        
        The 12 EGAD objects represent a wide diversity of geometries, including different sizes; feature types and thicknesses. The EGAD objects were all scaled to have a maximum width of \SI{55}{\milli\meter}. The three custom objects were generated as random splines, which were mirrored to make a symmetric object and then extruded, giving a set of amorphous soft shapes with distinct features.

        The objects were printed in three hardness levels: Shore 40A (softest), 60A and 85A (hardest), giving a range from much softer than the gripper's TPU material up to approximately the same hardness as the gripper.

\section{Experimental Results} \label{experiments}

        \begin{figure}[t]
                \centering
                \includegraphics[width=0.9\linewidth]{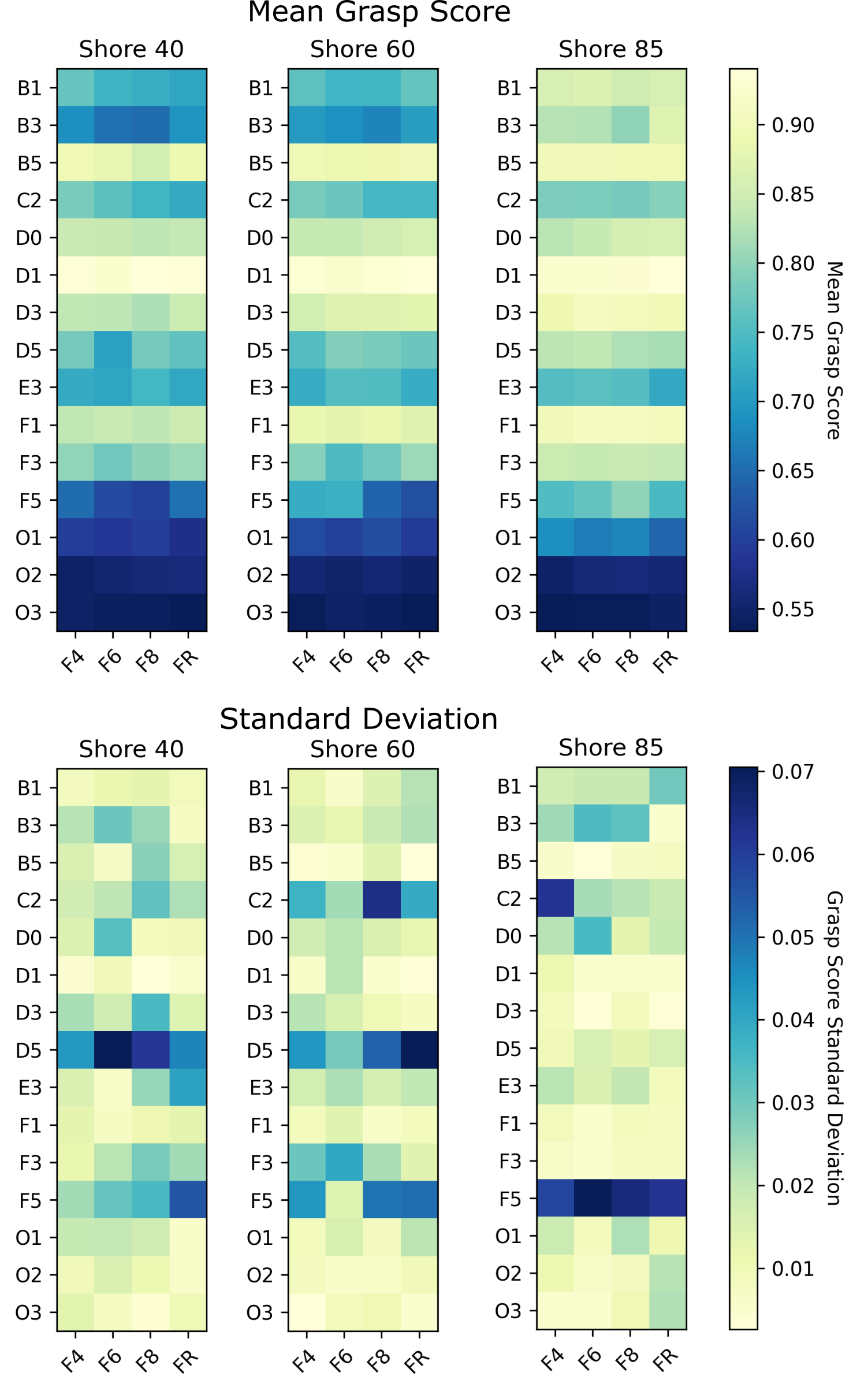}
                \caption{Heatmaps of the complete grasp evaluation dataset, showing the mean and standard deviations of each set of grasps.}
                \label{fig:heatmap}
            \end{figure}

    In this section we seek to answer the key questions: When is soft gripping important? What is a good gripper for soft objects? How well do soft grippers generalise across objects?
    To address these questions a total of 900 grasps were evaluated, comprising 15 object geometries, 3 object materials, 4 grippers, and 5 repeats of each. 
    
    The mean grasp score and standard deviation of each (geometry, material, gripper) pairing is presented in Figure~\ref{fig:heatmap}, indicating the grasp quality and repeatability, respectively.
    Every test in the 900 grasps was successful, with the objects able to be stably grasped and lifted, hence grasp scores are all between 0.5 and 1. The lowest score recorded was 0.517 (Object O3 Shore 40A with Rigid Grippers) and the highest was 0.940 (Object D1 Shore 85A with Rigid Grippers). 

\begin{figure}[t]
                \centering
                \includegraphics[width=0.8\linewidth]{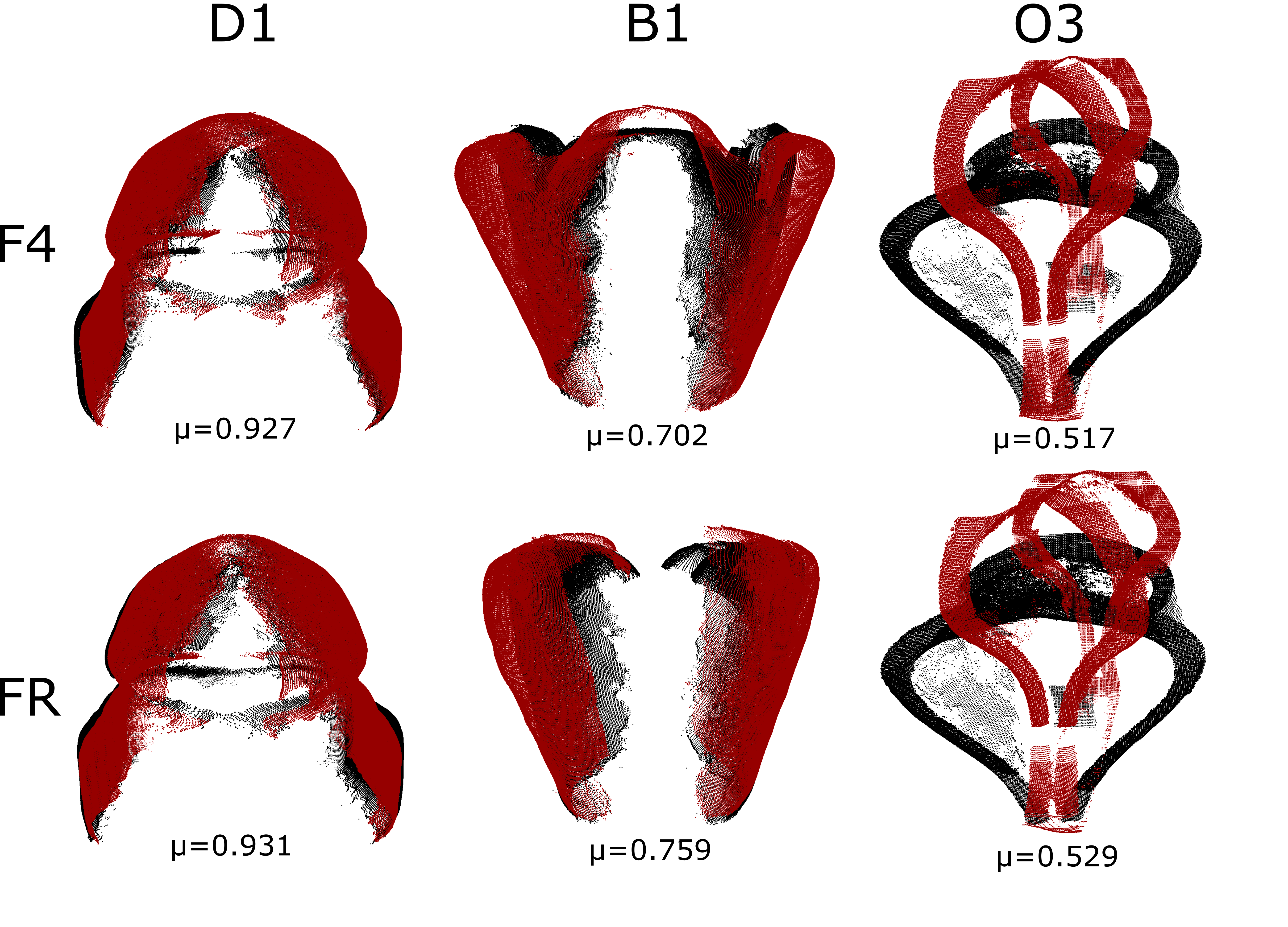}
                \caption{Point clouds of selected objects (all Shore 40A), comparing the 4-rib and Rigid Fin-Ray grasps on a relatively stiff (D1), intermediate (B1) and relatively soft (O3) object. The pre-grasp point cloud is black and the point cloud during grasping is red, with mean grasp scores for each object indicated.}
                \label{fig:pointclouds}
            \end{figure}

\begin{figure*}[t]
                \centering
                \includegraphics[width=\linewidth]{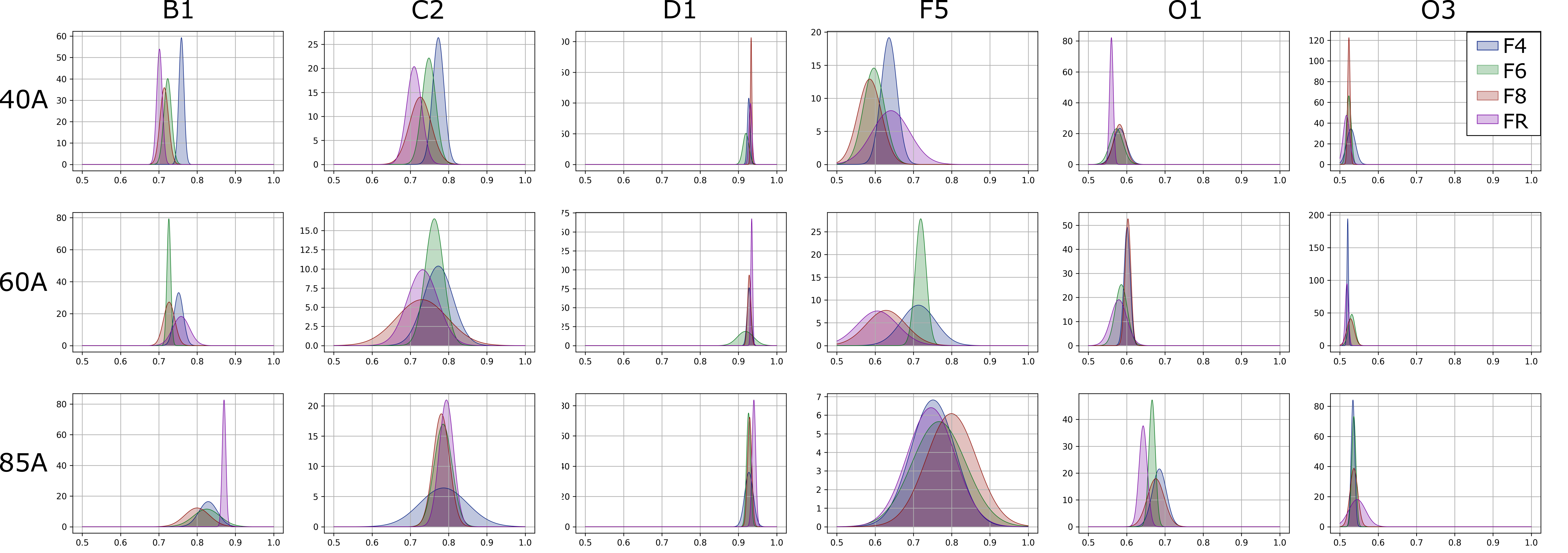}
                \caption{Score distributions for select objects, illustrating: Objects with unique rankings (B1 40A, C2 40A); a relatively rigid object (D1); an object with thin features causing large variation (F5); increasing trend of gripper scores across stiffnesses (B1, O1); hollow soft evaluation objects (O1, O3).}
                \label{fig:Combined_results}
            \end{figure*}

    The deformation experienced by these objects and an intermediate one are shown in Figure \ref{fig:pointclouds}. The evaluation objects, broadly, fall into three categories: (i) Those where the object is relatively stiff and little difference was observed between soft and rigid grippers, e.g. D1. (ii) Those which are very soft (even relative to the 4-rib Fin-Ray) and hence experience large deformations across all fingers (O2, O3). (iii) Those where the effective stiffness of the object was comparable to the grippers tested and exhibited clearly separable scores for each gripper within an object (e.g. B1 Shore 40A) or showing a clear trend across objects for the grippers collectively (e.g. B3). We discuss these in further detail below and provide quantitative results in Figure \ref{fig:Combined_results}.

    \subsection{Relatively Stiff Objects}
     The \textit{effective stiffness} of any structure is a function of its geometry, materials and the loading conditions (where and how forces are applied). Soft gripping is most valuable where the gripper's effective stiffness is similar (to an order of magnitude) to the object. The gripper being slightly softer than the object is beneficial as the gripper will absorb most of the strain energy produced during the grasp, preventing damage to the object, but if it is too soft it will be unable to support the object's weight. In contrast, if the gripper is much stiffer than the object, it performs as if it is rigid and the benefit of soft gripping is lost. The latter effect is seen in the objects D1 and B5, which both have a solid core of material, making them relatively stiff at all three shore hardnesses. The result is uniformly high grasping scores, without a meaningful performance difference between the soft and rigid grippers.

    \subsection{Relatively Soft}
        The three custom objects were computationally generated for both softness and symmetry. As such we expect low mean grasp scores and with little variation between them. 
        Experimentally these relatively soft objects experience such large deformations during grasping that the choice of gripper is immaterial, especially O2 and O3. Interestingly, the `corrugated' surface of O1 increased its effective stiffness and prevented the entire structure from flattening when grasped (as with O2 and O3). As a result, a larger range of scores occur across fingers and objects, and the stiffness of the rigid gripper can be clearly separated from the 3 Fin-Rays.

    \subsection{Just Right}

        Between these two extremes of objects which are too soft or too firm for soft grasping to be valuable, there is a set of objects which are ideally suited to distinguishing the performance of the evaluated grippers. This can be evaluated both within objects and across objects. Within objects, we expect higher grasp scores from softer grippers; across objects we expect all grippers to score better on harder objects than softer ones. Whilst no object perfectly displayed both characteristics, several showed one or the other. For example, going by mean grasp scores, B1 Shore 40A, C2 Shore 40A, and O3 Shore 40A, all rank the grippers from softest to hardest. However, for the same objects printed at Shore 85A hardness, the scoring curves converge such that they are unable to meaningfully separate the grippers. This suggests soft grasping is beneficial in the softer objects, but at the harder ones it is of low benefit within the stiffness range of the grippers evaluated. More compliant/lower stiffness grippers are required for these objects, these could be Fin-Rays with fewer ribs or softer material, or a different design altogether.

        For objects B1, D3, O1 and O3, the four sets of fingers collectively exhibit better performance on the higher Shore values compared to the lower ones, indicating the object's stiffness materially contributes to grasp score. Whilst the softer grippers typically outperformed the harder ones in these objects, it was not universal. In some objects, small changes in the object position relative to the gripper caused substantially different grasp behaviour, which can manifest as a large standard deviation in grasp scores for a particular object-gripper pair or an unexpected gripper ranking, both of which are present in D3. This is more generally the case in objects with thin features (e.g. C2, D5, F5), which gave highly varied grasp qualities and had large standard deviations as a result. Such objects would benefit from a larger sample set.
                  
\section{Discussion and Conclusion} \label{section: conclusion} \label{conclusion}
    This work proposes SoGraB as a new benchmarking protocol for soft grasp evaluation, which captures object deformation as a proxy for stress. As it does not rely on any specific gripper or object instrumentation, it readily generalises to any gripper-object pairing (excluding those where the object is completely occluded). Using SoGraB we produce a baseline dataset comprising 900 grasps across 45 objects and 4 grippers. Through it we show there is a defined range of stiffnesses in which soft grippers thrive, outside of this range there is no quantifiable difference between soft and rigid grippers. As this range is specific to each object and gripper, we demonstrate that SoGraB is an effective method to evaluate soft grasping performance. The protocol is designed to be implemented in any robotics lab with commonly available hardware (robot arm, 3D printer and 3D camara). Future users can contribute to the dataset by running the SoGraB protocol by: expanded the range of objects to evaluate new shapes and hardnesses, or benchmark new soft grippers against the existing object dataset. In selecting objects, we recommend symmetric objects without thin external features, as these give repeatable data. For grippers, any design can be evaluated so long as the object remains at least partially visible.
    Through the ongoing use of SoGraB we aim to improve the quality of soft gripper designs, and identify the most valuable use cases for soft gripping. We believe SoGraB  is a valuable benchmark to compare against existing and future designs, and hope it will help improve the quality of future soft gripper designs. 
    
\section*{Acknowledgment}
This work is supported by the Science and Industry Endowment Fund (SIEF) and CSIRO’s Future Digital Manufacturing Fund.

\bibliographystyle{IEEEtran}

\bibliography{bib}

\begin{thebibliography}{10}
\providecommand{\url}[1]{#1}
\csname url@samestyle\endcsname
\providecommand{\newblock}{\relax}
\providecommand{\bibinfo}[2]{#2}
\providecommand{\BIBentrySTDinterwordspacing}{\spaceskip=0pt\relax}
\providecommand{\BIBentryALTinterwordstretchfactor}{4}
\providecommand{\BIBentryALTinterwordspacing}{\spaceskip=\fontdimen2\font plus
\BIBentryALTinterwordstretchfactor\fontdimen3\font minus \fontdimen4\font\relax}
\providecommand{\BIBforeignlanguage}[2]{{%
\expandafter\ifx\csname l@#1\endcsname\relax
\typeout{** WARNING: IEEEtran.bst: No hyphenation pattern has been}%
\typeout{** loaded for the language `#1'. Using the pattern for}%
\typeout{** the default language instead.}%
\else
\language=\csname l@#1\endcsname
\fi
#2}}
\providecommand{\BIBdecl}{\relax}
\BIBdecl

\bibitem{Pinskier2024}
R.~Kanno, P.~H. Nguyen, J.~Pinskier, D.~Howard, S.~Song, and M.~Kovac, ``{Hybrid Soft Electrostatic Metamaterial Gripper for Multi-surface, Multi-object Adaptation},'' \emph{2024 IEEE 7th International Conference on Soft Robotics, RoboSoft 2024}, pp. 851--857, 2024.

\bibitem{Ilievski2011}
F.~Ilievski, A.~D. Mazzeo, R.~F. Shepherd, X.~Chen, and G.~M. Whitesides, ``{Soft Robotics for Chemists},'' \emph{Angewandte Chemie}, vol. 123, no.~8, pp. 1930--1935, 2011.

\bibitem{Smith2023}
L.~Smith and R.~MacCurdy, ``{SoRoForge: End-to-End Soft Actuator Design},'' \emph{IEEE Transactions on Automation Science and Engineering}, vol.~20, no.~3, pp. 1475--1486, 2023.

\bibitem{Pinskier2024a}
J.~Pinskier, J.~Brett, and D.~Howard, ``{Towards Bespoke Soft Grippers through Voxel-Scale Metamaterial Topology Optimisation},'' in \emph{Robosoft 2024}.\hskip 1em plus 0.5em minus 0.4em\relax San Diego: IEEE, 2024, pp. 1--8.

\bibitem{Pinskier2024c}
J.~Pinskier, X.~Wang, L.~Liow, Y.~Xie, P.~Kumar, M.~Langelaar, and D.~Howard, ``{Diversity-Based Topology Optimization of Soft Robotic Grippers},'' \emph{Advanced Intelligent Systems}, vol.~6, no.~4, 2024.

\bibitem{Xie2024}
Y.~Xie, J.~Pinskier, X.~Wang, and D.~Howard, ``{Evolutionary Seeding of Diverse Structural Design Solutions via Topology Optimization},'' \emph{ACM Transactions on Evolutionary Learning and Optimization}, 2024.

\bibitem{Baines2024}
R.~Baines, D.~Shah, J.~Marvel, J.~Case, and A.~Spielberg, ``{The need for reproducible research in soft robotics},'' \emph{Nature Machine Intelligence}, vol.~6, no.~7, pp. 740--741, 2024.

\bibitem{doi:10.1126/scirobotics.abg6049}
E.~W. Hawkes, C.~Majidi, and M.~T. Tolley, ``Hard questions for soft robotics,'' \emph{Science Robotics}, vol.~6, no.~53, p. eabg6049, 2021.

\bibitem{Zimmer2019}
J.~Zimmer, T.~Hellebrekers, T.~Asfour, C.~Majidi, and O.~Kroemer, ``{Predicting Grasp Success with a Soft Sensing Skin and Shape-Memory Actuated Gripper},'' \emph{IEEE International Conference on Intelligent Robots and Systems}, pp. 7120--7127, 2019.

\bibitem{10122060}
D.~Howard, J.~O'Connor, J.~Letchford, T.~Joseph, S.~Lin, S.~Baldwin, and G.~Delaney, ``A comprehensive dataset of grains for granular jamming in soft robotics: Grip strength and shock absorption,'' in \emph{2023 IEEE International Conference on Soft Robotics (RoboSoft)}, 2023, pp. 1--8.

\bibitem{young2001roark}
W.~Young and R.~Budynas, \emph{Roark's Formulas for Stress and Strain}, ser. MacGraw-Hill international edition.\hskip 1em plus 0.5em minus 0.4em\relax McGraw Hill LLC, 2001.

\bibitem{Low2021}
J.~H. Low, P.~M. Khin, Q.~Q. Han, H.~Yao, Y.~S. Teoh, Y.~Zeng, S.~Li, J.~Liu, Z.~Liu, P.~{Valdivia y Alvarado}, I.-M. Chen, B.~C.~K. Tee, and C.~H. Yeow, ``{Sensorized Reconfigurable Soft Robotic Gripper System for Automated Food Handling},'' \emph{IEEE/ASME Transactions on Mechatronics}, pp. 1--12, 2021.

\bibitem{10538419}
M.~Knopke, L.~Zhu, P.~Corke, and F.~Zhang, ``Towards assessing compliant robotic grasping from first-object perspective via instrumented objects,'' \emph{IEEE Robotics and Automation Letters}, vol.~9, no.~7, pp. 6320--6327, 2024.

\bibitem{Junge2022}
K.~Junge and J.~Hughes, ``{Soft Sensorized Physical Twin for Harvesting Raspberries},'' \emph{2022 IEEE 5th International Conference on Soft Robotics, RoboSoft 2022}, pp. 601--606, 2022.

\bibitem{Falco2020}
J.~Falco, D.~Hemphill, K.~Kimble, E.~Messina, A.~Norton, R.~Ropelato, and H.~Yanco, ``{Benchmarking Protocols for Evaluating Grasp Strength, Grasp Cycle Time, Finger Strength, and Finger Repeatability of Robot End-Effectors},'' \emph{IEEE Robotics and Automation Letters}, vol.~5, no.~2, pp. 644--651, 2020.

\bibitem{mahler_guest_2018}
J.~Mahler, R.~Platt, A.~Rodriguez, M.~Ciocarlie, A.~Dollar, R.~Detry, M.~A. Roa, H.~Yanco, A.~Norton, J.~Falco, K.~v. Wyk, E.~Messina, J.~â. Leitner, D.~Morrison, M.~Mason, O.~Brock, L.~Odhner, A.~Kurenkov, M.~Matl, and K.~Goldberg, ``Guest {Editorial} {Open} {Discussion} of {Robot} {Grasping} {Benchmarks}, {Protocols}, and {Metrics},'' \emph{IEEE Transactions on Automation Science and Engineering}, vol.~15, no.~4, pp. 1440--1442, 2018, number: 4.

\bibitem{Rubert2018}
C.~Rubert, B.~Le{\'{o}}n, A.~Morales, and J.~Sancho-Bru, ``{Characterisation of Grasp Quality Metrics},'' \emph{Journal of Intelligent and Robotic Systems: Theory and Applications}, vol.~89, no. 3-4, pp. 319--342, 2018.

\bibitem{roa_grasp_2015}
M.~A. Roa and R.~Suárez, ``\BIBforeignlanguage{en}{Grasp quality measures: review and performance},'' \emph{\BIBforeignlanguage{en}{Autonomous Robots}}, vol.~38, no.~1, pp. 65--88, Jan. 2015, number: 1.

\bibitem{pozzi_grasp_2017}
M.~Pozzi, M.~Malvezzi, and D.~Prattichizzo, ``On {Grasp} {Quality} {Measures}: {Grasp} {Robustness} and {Contact} {Force} {Distribution} in {Underactuated} and {Compliant} {Robotic} {Hands},'' \emph{IEEE Robotics and Automation Letters}, vol.~2, no.~1, pp. 329--336, Jan. 2017, number: 1 Conference Name: IEEE Robotics and Automation Letters.

\bibitem{sotiropoulos_benchmarking_2018}
P.~Sotiropoulos, M.~A. Roa, M.~F. Martins, W.~Fried, H.~Mnyusiwalla, P.~Triantafyllou, and G.~Deacon, ``A {Benchmarking} {Framework} for {Systematic} {Evaluation} of {Compliant} {Under}-{Actuated} {Soft} {End} {Effectors} in an {Industrial} {Context},'' in \emph{2018 {IEEE}-{RAS} 18th {International} {Conference} on {Humanoid} {Robots} ({Humanoids})}, Nov. 2018, pp. 280--283, iSSN: 2164-0580.

\bibitem{murrilo_multigrippergrasp_2024}
L.~F.~C. Murrilo, N.~Khargonkar, B.~Prabhakaran, and Y.~Xiang, ``{MultiGripperGrasp}: {A} {Dataset} for {Robotic} {Grasping} from {Parallel} {Jaw} {Grippers} to {Dexterous} {Hands},'' Mar. 2024, issue: arXiv:2403.09841 arXiv:2403.09841 [cs].

\bibitem{calli_ycb_2015}
B.~Calli, A.~Singh, A.~Walsman, S.~Srinivasa, P.~Abbeel, and A.~M. Dollar, ``The {YCB} object and {Model} set: {Towards} common benchmarks for manipulation research,'' in \emph{2015 {International} {Conference} on {Advanced} {Robotics} ({ICAR})}, 2015, pp. 510--517.

\bibitem{mahler_dex-net_2016}
J.~Mahler, F.~T. Pokorny, B.~Hou, M.~Roderick, M.~Laskey, M.~Aubry, K.~Kohlhoff, T.~Kröger, J.~Kuffner, and K.~Goldberg, ``Dex-{Net} 1.0: {A} cloud-based network of {3D} objects for robust grasp planning using a {Multi}-{Armed} {Bandit} model with correlated rewards,'' in \emph{2016 {IEEE} {International} {Conference} on {Robotics} and {Automation} ({ICRA})}, 2016, pp. 1957--1964.

\bibitem{morrison_egad_2020}
D.~Morrison, P.~Corke, and J.~Leitner, ``{EGAD}! {An} {Evolved} {Grasping} {Analysis} {Dataset} for {Diversity} and {Reproducibility} in {Robotic} {Manipulation},'' \emph{IEEE Robotics and Automation Letters}, vol.~5, no.~3, pp. 4368--4375, 2020, number: 3.

\bibitem{wu_chamfer_2021}
T.~Wu, L.~Pan, J.~Zhang, T.~WANG, Z.~Liu, and D.~Lin, ``Chamfer {Distance} as a {Comprehensive} {Metric} for {Point} {Cloud} {Completion},'' in \emph{Advances in {Neural} {Information} {Processing} {Systems}}, vol.~34.\hskip 1em plus 0.5em minus 0.4em\relax Curran Associates, Inc., 2021, pp. 29\,088--29\,100.

\end{thebibliography}

\end{document}